\documentclass[lettersize,journal]{IEEEtran}
\usepackage{amsmath,amsfonts}
\usepackage{algorithmic}
\usepackage{algorithm}
\usepackage{array}
\usepackage[caption=false,font=normalsize,labelfont=sf,textfont=sf]{subfig}
\usepackage{textcomp}
\usepackage{stfloats}
\usepackage{url}
\usepackage{verbatim}
\usepackage{graphicx}
\usepackage{booktabs}
\usepackage{multirow}
\usepackage{multicol}
\begin{document}

\title{A Two-stage Personalized Virtual Try-on \\Framework with Shape Control and Texture Guidance}

\author{Shufang~Zhang,~\IEEEmembership{Member,~IEEE,}
        Minxue~Ni,
        Lei~Wang,
        Wenxin~Ding,
        Shuai~Chen,
        and~Yuhong~Liu,~\IEEEmembership{Senior Member,~IEEE}
        }



\maketitle

\begin{abstract}
The Diffusion model has a strong ability to generate wild images. However, the model can just generate inaccurate images with the guidance of text, which makes it very challenging to directly apply the text-guided generative model for virtual try-on scenarios. Taking images as guiding conditions of the diffusion model, this paper proposes a brand new personalized virtual try-on model (PE-VITON), which uses the two stages (shape control and texture guidance) to decouple the clothing attributes. Specifically, the proposed model adaptively matches the clothing to human body parts through the Shape Control Module (SCM) to mitigate the misalignment of the clothing and the human body parts. The semantic information of the input clothing is parsed by the Texture Guided Module (TGM), and the corresponding texture is generated by directional guidance. Therefore, this model can effectively solve the problems of weak reduction of clothing folds, poor generation effect under complex human posture, blurred edges of clothing, and unclear texture styles in traditional try-on methods. Meanwhile, the model can automatically enhance the generated clothing folds and textures according to the human posture, and improve the authenticity of virtual try-on. In this paper, qualitative and quantitative experiments are carried out on high-resolution paired and unpaired datasets, the results show that the proposed model outperforms the state-of-the-art model.
\end{abstract}


\section{Introduction}
\IEEEPARstart{R}{ecently}, Denoising Diffusion Probabilistic Model (DDPM)\cite{ho2020denoising} and score-based generative models\cite{song2019generative,song2020score} have shown great capabilities in image generation tasks. Compared with traditional generative models based on GAN\cite{he2022style,sarkar2021humangan,lewis2021tryongan,fu2022stylegan} and VAE\cite{jiang2022text2human}, these models can achieve the generation of a series of realistic style images, which have been widely recognized in various fields\cite{ho2020denoising,song2020denoising}.

\begin{figure}[!t]
\centering
\includegraphics[scale=0.117]{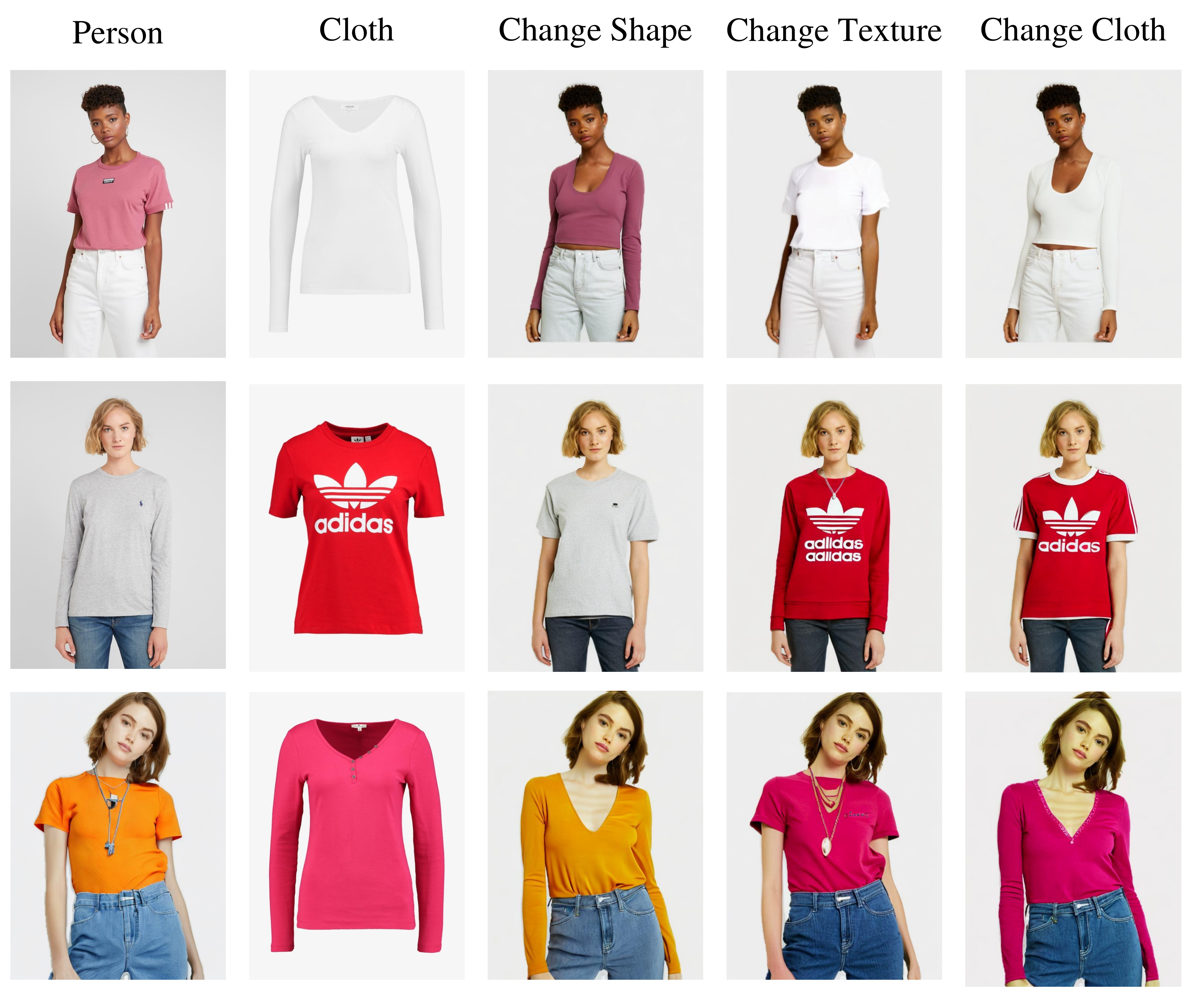}
\caption{\textbf{Personalized try-on results generated by our method.} Each row represents the result of controlling the shape and texture of the garment with the clothing condition. The last column shows the result of wearing the target garment. Our model enables decoupling control of the garment's shape and texture.}
\label{fig: first show}
\end{figure}

In the field of virtual try-on, researchers have tried to use text as guidance for DDPM models. Zhang et al. \cite{zhang2022humandiffusion} fine-tuned clothing styles based on input text information. Pernuš et al. \cite{pernuvs2023fice} generated random faces and costumes based on input text. Bhunia et al. \cite{bhunia2023person} achieved pose-guided human generation by developing a texture diffusion block based on cross-attention, and a texture diffusion block based on a multi-scale texture pattern encoding source image. The generation results of the above method are very realistic. However, due to the limited guidance ability of simple text description or pose information, the provided guidance information is ambiguous. Hence, the freedom of the generated results is enormous, which makes it very difficult to obtain accurate guidance, and limits the effectiveness of the model to obtain satisfactory generated results for users.

In response to the limited ability of text guidance, many researchers began to explore the possibility of image guidance. The researchers parsed the image information into semantic information \cite{yang2023paint} or style features \cite{jeong2023training}, and injected it into the diffusion model as conditions, so as to guide the generation of the target result map on the basis of the source map. Based on these heuristics, this paper takes images as the guiding condition of the diffusion model and proposes a generation try-on guided by clothing images. Specifically, the proposed scheme can generate personalized virtual try-on by transforming the target clothing image into latent semantic information and guiding the corresponding clothing position of the source image. The model consists of a Shape Control Module (SCM) and a Texture Guided Module (TGM). The shape control module aims to generate segmentation results that are consistent with human bodies by controlling and editing clothing segmentation, matching body parts with target clothing, adjusting segmentation information according to the target clothing style adaptively, determining non-target body areas and occluded body areas dynamically, and adjusting semantic layout in an orderly manner. The texture guided module takes the diffusion model as an image prior, transforms the provided clothing style into the embedding of potential space as the conditional guidance, and applies the powerful generation ability of high-quality wild images of the model to the real scene of personalized virtual try-on.

The proposed model effectively solves some major limitations of traditional virtual try-on models. For example, traditional try-on models can only transfer simple clothing shape and texture \cite{xie2022pasta,xie2021towards,cui2021dressing,men2020controllable}, but not truly restore the fold style of clothing. Furthermore, the clothing shapes are largely limited by the original clothing shapes, and the junctions between the clothing and the human body are easily blurred. Currently, most virtual try-on methods can only be used for paired scenes and trained on the paired datasets\cite{zhang2023limb},i.e., a person image and its corresponding in-shop garment. This means the training stage depends on the laborious data-collection processes. Furthermore, only when the in-shop tiled garment is obtained can users utilize their own pictures for virtual try-on in the application. A few attempts have been made towards unpaired virtual try-on, that is, the clothes worn in the model picture are going to be transferred to the target body picture. These methods have their own practical applications, but they also have drawbacks. For example, in the paired scene, it is difficult for users to get high-quality in-shop tiled garments. The unpaired method makes it difficult to decouple the poses of two pictures and achieve garment distortion. Therefore, in this study, we proposed a virtual try-on scheme that can address these issues and reflect personalized characteristics, so that: (1) the powerful generation ability of DDPM can adaptively adjust the folds of the generated clothing according to the posture of the user, making the wearing effect more natural; (2) texture enhancement can be achieved, so that the generated texture style is clearer and more distinctive; (3) users are allowed to alter the shape and texture style of the generated clothing through different in-shop tiled garments or model images guidance, which meets the diversified try-on needs in real scenes; (4) it can be adapted to both paired and unpaired scenarios simultaneously. This method breaks through the boundaries of the traditional virtual try-on method and overcomes the dependence of traditional try-on methods on clothing distortion function and posture style decoupling. Furthermore, by solving the problem of mismatch between the human body and clothing, and generating more realistic and diverse try-on results, the proposed scheme can deal with more complex try-on scenarios and needs.

Comprehensive experiments have been carried out on the DeepFashion Multimodal dataset and VITON-HD dataset, and the results prove the superiority of the proposed method when compared with state-of-the-art methods. The main contributions of this paper can be summarized as follows:

\begin{itemize}
\item We propose a new two-stage personalized virtual try-on framework, which includes a shape control module and a texture-guided module. By dynamically changing the shape and style of the target garment through the cascading method, the proposed scheme can achieve fine-grained control and adjustment on the generated results, so as to realize personalized virtual try-on.
\item We use generative models to replace the unnatural structure of the warped module in the traditional try-on model. The generative model converts the clothing information into semantic information in the latent space, and guides the model to understand the clothing context information, so as to generate realistic try-on results. Therefore, this method avoids a series of problems caused by traditional warping methods, such as mismatch between body parts and clothing, difficulty in learning complex postures, and body parts covering clothing.
\item We break the limitations of paired and unpaired dataset
 and achieve state-of-the-art performance on the high-resolution\cite{xu2021virtual} datasets. For the SCM and TGM, their garment input can be either the in-store tiled garment image or the model image, which does not affect the final result. We use both paired and unpaired datasets as the training set of the model, and our model achieves good generation results in both application scenarios.
\end{itemize}

\section{Related Word}

\begin{figure*}[h]
    \includegraphics[scale=0.137]{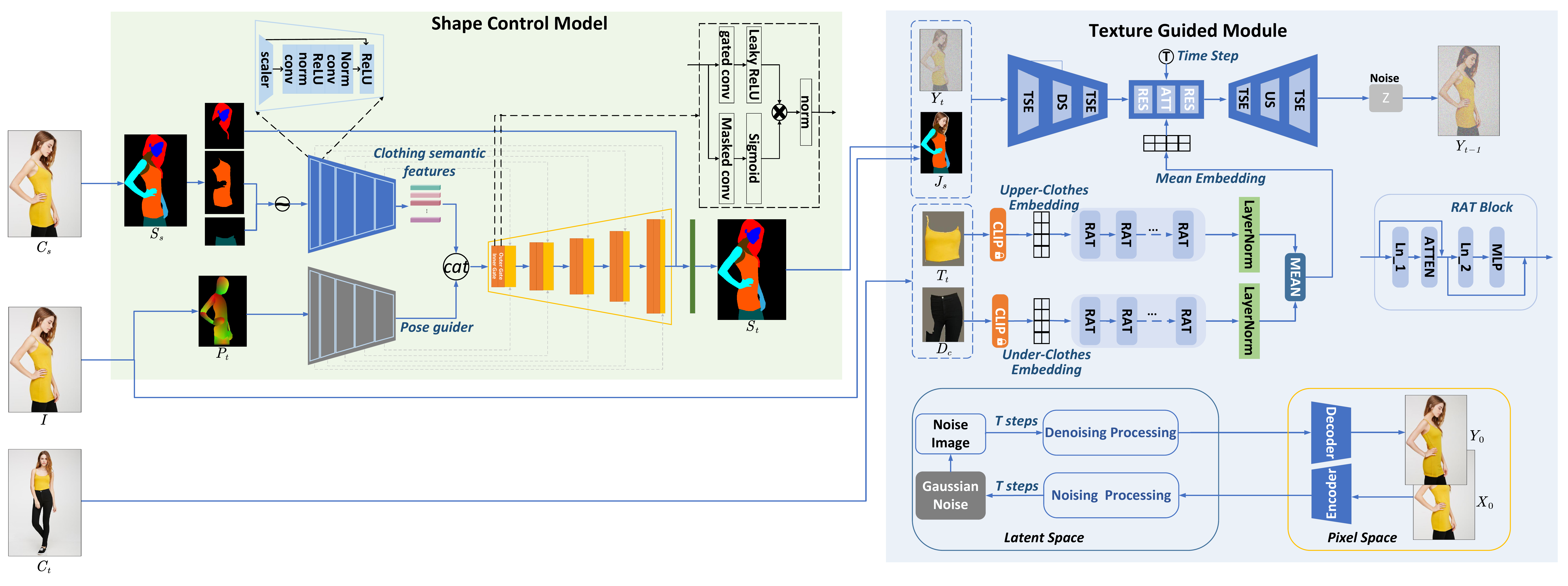}
    \captionsetup{justification=justified}
    \caption{\textbf{Overview of the proposed framework (PE-VITON)}. We use $I$, $C_s$ and $C_t$ to represent the human source image, shape control condition and texture control condition respectively. During the training process, the $C_s$ and the $C_t$ are the same image. In the SCM stage, segmentation $S_s$ and pose information $P_t$ are used to obtain the target segmentation $S_t$. In the TGM stage, we stitch the $S_t$ and the human source image $I$ to get $J_s$, which helps to retain the head information. The garment obtained from $C_t$ is encoded to embedding as the conditional control to generate the final result.}
    \label{fig:pipeline}
\end{figure*}

\subsection{Preliminaries}
Diffusion probability models are a class of latent variable models consisting of forward diffusion processes and reverse diffusion processes. The forward process is a Markov chain in which noise accumulates into the picture through T operations. Each step be described as:
\begin{equation} 
 x_t = \sqrt{\alpha_t}x_{t-1}+\sqrt{1-\alpha_t}\varepsilon_{t-1}
\end{equation} where $\varepsilon_{t-1}$represents the noise, $\alpha_t = 1 - \beta_t$, $\beta_t$ is the hyperparameter of the diffusion model, $x_{t-1}$ represents the last step image, $x_t$ represents the current image.

In the forward process, as $t$ gradually increases, noise is added to the image step by step, $x_t$ is closer to pure noise. The backward process is the denoising inference process of the diffusion model, that is, the random noise is gradually reduced to the original picture by predicting the noise. The reverse process can be described as:
\begin{equation} 
 x_{t-1} = \frac{1}{\sqrt{a_t}}(x_t-\frac{1-\alpha_t}{1-\overline{\alpha_t}}\varepsilon_\theta(x_t,t))+\sigma_tz
\end{equation} where in every step, use $x_t$ to subtract the currently predicted noise $\frac{1-\alpha_t}{1-\overline{\alpha_t}}\varepsilon_\theta(x_t,t)$, and add a small noise $\sigma_tz$ for correction, that is, the image with noise removed in this step is obtained.

\subsection{Conditional image generation based on the diffusion model}
With the advent of the diffusion model, the image generation task has made a huge breakthrough. With the ability to preserve the semantic structure of data, diffusion models can achieve higher diversity and fidelity than GAN-based generative models. GLIDE\cite{nichol2021glide} and DALLE-2\cite{ramesh2022hierarchical} propose to generate novel images based on the text descriptions provided by humans, which have never been seen in the dataset before. Stable Diffusion\cite{rombach2022high} generates images by iterating over “denoised” data in a latent representation space, enabling text generation tasks to be completed on consumer GPUs. Retrieval-Diffusion\cite{blattmann2022retrieval} introduces nearest-neighbor retrieval to improve generation quality and diversity. Humandiffusion\cite{zhang2022humandiffusion} controls the results of human image generation through text input on the basis of segmentation.

Many researchers have proposed text-guided conditional image generation using diffusion models. DiffusionCLIP\cite{kim2022diffusionclip} fine-tunes images by combining pre-trained diffusion models with CLIPloss. Dreambooth\cite{ruiz2023dreambooth} implements directional modification of image properties through text. Imagic\cite{kawar2023imagic} uses complex language input to achieve multi-attribute adjustment of images. Blended Diffusion\cite{avrahami2022blended} proposes a multi-step mixing process that utilizes user-supplied masks for local manipulation. FICE\cite{pernuvs2023fice} adjusts the clothing style of a human body image by converting the entered text information into style attributes.

Although the use of text can achieve simple control over image generation and editing, text guidance still has its disadvantages of lack of detailed information and low accuracy. On the other hand, image guidance can provide more detailed and accurate information. As a result, some work begin to emerge exploring the possibility of image-guided generation. BBDM\cite{li2023bbdm} and DeepPortraitDrawing\cite{wu2023deepportraitdrawing} use sketches as guiding conditions to generate corresponding images. Many studies have been proposed for image restoration tasks. For example, MCG\cite{chung2022improving} and RePaint\cite{lugmayr2022repaint} realize image repair and restoration for image occlusion. Palette\cite{saharia2022palette} proposes a unified framework for image-to-image conversion based on a conditional diffusion model for conversion tasks (i.e., coloring, fixing, uncropping, and JPEG recovery).

In addition to using image guidance to generate new images or realize image-to-image conversion, the researchers also explored the possibility of using images as conditions to guide other image generation. For example, Paint by Example\cite{yang2023paint} uses input images as the condition to guide the image mask part to generate relevant semantic content; DiffStyle\cite{jeong2023training} realizes style injection through image input, thereby realizing training-free feed-forward style transfer; \cite{gal2022image} proposes the concept of “textual inversion”, that is, pseudo-words are generated from the input image, and are embedded in the new scene to generate relevant image results. These methods directly use images as guiding conditions to guide the source image to generate relevant brand-new images, which proves the feasibility of image guidance.

\subsection{Image-based virtual try-on}
The image-based virtual try-on task is designed to generate realistic try-on results based on the given garment image to be fitted. According to the source of clothing images, virtual try-on methods can be mainly divided into paired-image methods and unpaired-image methods. The ``paired" means the dataset consists of the pairs of model images and the corresponding in-shop tiled garment images. The model image in it is usually a partial image (such as an upper body image or lower body image). While the ``unpaired" means the dataset only contains the model images and does not provide the corresponding tiled garment images. The current paired methods\cite{lee2022towards,ge2021disentangled,ge2021parser,han2018viton,minar2020cp,yang2020towards,han2019clothflow,yu2019vtnfp,xing2022virtual}, by following the paradigm from clothing to human body, use image pairs of in-shop tiled garment and body image to transfer the in-shop garment to the target human body. Thin plate spline (TPS) transformation method\cite{belongie2002shape,hu2022spg} used to warp the garment can easily lead to undesirable distortion results when the human pose is too complex. Using appearance flow\cite{yang2023occlumix} can address this problem to a certain extent. However, it cannot fully address the remaining artifacts in the generated results. In addition, both of these methods struggle to generate non-blurry clothing textures and folds.

Since it is difficult to collect paired datasets and to acquire in-shop tiled garment images for try-on tasks, some researchers have tried to use unpaired datasets to deal with virtual try-on tasks\cite{lewis2021tryongan, xie2022pasta, xie2021towards, ge2021parser, neuberger2020image}. Although the current unpaired try-on methods can realize the simple transfer of clothing, they may lead to structural deformation of complex style clothing and human pose. Furthermore, the unpaired methods mainly target low-resolution datasets instead of high-resolution datasets. Compared with the paired-image virtual try-on methods, the unpaired virtual try-on methods are still at their early stages, so it is difficult to achieve good try-on results using unpaired methods.

Paired methods and unpaired methods both have their advantages and disadvantages. Actually, Dividing tasks through the pairing of datasets limits the application of the model, so these methods have relatively high requirements for the image input. In order to bridge the gap between these two categories of methods, this paper proposes a more flexible try-on method based on the diffusion model, which breaks the limitation of paired and unpaired methods and makes it more convenient to facilitate users for virtual try-on tasks.

\section{Method}
The training stage pipeline of PE-VITON proposed in this paper is shown in Fig. \ref{fig:pipeline}, which contains two parts: a Shape Control Module (SCM) and a Texture Guided Module (TGM). The input of the model includes a human source image $I\in\mathbb{R}^{3\times H\times W}$, a target shape garment image $C_s\in\mathbb{R}^{3\times H\times W}$, and a target texture garment $C_t\in\mathbb{R}^{3\times H\times W}$, where $H$ and $W$ represent the height and width of the clothing respectively. The ultimate goal is to generate an image of $Y_0\in\mathbb{R}^{3\times H\times W}$, which represents the generated clothing with the shape of $C_s$ and texture style of $C_t$, and the human posture and shape in the generated images remain consistent with $I$. In the training stage, the target shape clothing and the target texture clothing are consistent in terms of the shape and texture of the clothing with the clothing of the human source image, which means they are the same garment. The loss is calculated as the offset between the segmentation obtained in the shape control stage and the segmentation $S_s$ of the human source image; as well as the offset between the result of the texture generation stage and the human source image.

Considering the human source representation $I$ and the target shape of the clothing $C_s$, SCM deforms $C_s$, generates a new segmentation map, and adaptively matches the target clothing to the corresponding position of the human posture. Since vanilla convolution tends to treat all pixels as valid information, which leads to poor segmentation results. Therefore, we propose to deconstruct the position information and shape information of clothing, and adopt gated convolution to filter out poor segmentation results, which allows the model to learn the shape of the specified clothing position (Section A). Then TGM relocates the SCM output results and clothing texture conditions to the latent space, and then generates reasonable try-on results according to the clothing shape and texture conditions, so as to achieve a smooth transition between the human body and clothing boundaries (Section B).

\subsection{Shape Control Module}
The architecture of the SCM is shown in Fig. \ref{fig:pipeline}, which consists of two encoders and a decoder. In this stage, we aim to generate the segmentation map $S_t$ of the person wearing the target shape clothing item $C_s$ and adaptively deform $C_s$ to fit the body pose of the source person $I$. During the training process, the densepose $P_t$ of the human body $I$ is leveraged as the auxiliary condition to help SCM learn pose information. The generated segmentation map $S_t$ can then serve as the input of TGM. Given source segmentation map $S_c$ and densepose $P_t$, the destruction function is employed first to get the clothing segmentation from $S_c$, and then an upper clothing or an under clothing will be chosen by the condition selector. Afterward, the feature pyramid is extracted from each encoder, which will be fed into the decoder, and aligned by the gated convolution. $G_{x,y}$ is used to align the features through the output of the last processing layer to obtain the target segmentation map $S_t$.

We add gated convolution to the front of each layer of the decoder (shown in Fig. \ref{fig:pipeline}) to avoid the disadvantage of vanilla convolution, where all pixels are considered valid information. Gated convolution can be expressed as:
\begin{align}
    & Gating_{y,x} = \sum\sum w_g\cdot C \notag\\
    & Feature_{y,x} = \sum\sum w_f \cdot C \\
    & G_{x,y} = lr(Feature_{y,x}) \cdot \sigma (Gating_{y,x}) \notag
\end{align}

where $\sigma$ represents the sigmoid function that keeps the output gating values between 0 and 1. $lr$ represents the LeakyReLU activation function. $w_g$ and $w_f$ represent the convolution kernel of gating and feature function respectively. $\cdot$ represents the element-wise multiplication. Gated convolution can learn the dynamic feature selection mechanism of each channel and spatial location, and take the semantic segmentation of some channels into consideration, which helps learn the mask position and feature information highlighted in different channels and repair the results. The model can inpaint the boundary between the distorted clothing and the human body from high to low level by adding gated convolution to each layer of the decoder, which helps the model generate clear clothing boundaries without semantic conflicts.

\subsection{Texture Guided Module}
In this stage, we aim to generate personalized try-on results $Y_0$ wearing a garment with a texture similar to the provided garment conditions $C_t$.

\textbf{Data processing}. During the training process, $x_0$ is migrated to the latent space through encoder $\varepsilon$, and then noises are added in $t$ steps to gradually obtain the noise image $Y_t$. Furthermore, the stitching input $J_s$ is obtained by stitching the segmentation $S_t$ with the head of the target human body so that the identity information of the try-on human can be retained. The process can be shown as:

\begin{scriptsize}
    \begin{align} 
        J_s = [I \cdot M_{k} + S \cdot (1-M_{k})] \cdot (1-M_{up}) \cdot (1-M_{un}) + 0.5 \cdot M_{un}
    \end{align}
\end{scriptsize}where $I$ represents the target human image; $S$ represents the generation segmentation $S_t$; $M_{k}$, $M_{up}$, and $M_{un}$ denote the head mask, the upper clothes mask, and the underclothes mask, respectively. Since there is a pixel-level misalignment between the segmentation result and the original image, which can cause an unsatisfactory white border in the generated result. To solve this problem, we propose an edge-enhancing information processing method that re-identifies and reassigns the label of the boundary through a pixel discriminator. The pixel discriminator is represented as:

\begin{footnotesize}
    \begin{equation} 
    label(J_s) = for \, label(i) \, in \, crop(J_s)0?\,label(neigh):label(i)
    \end{equation}
\end{footnotesize}where the $crop()$ function is used to get the $J_s$ result without background information; $neigh$ denotes the $N_s(p)$ range of the current pixel; and the $label(neigh)$ represents the mode of labels in the neighborhood.

\textbf{Data augmentation}. Considering that the user-input image may encounter low-quality problems such as poor lighting, color temperature deviation, and blurred edges, a series of data augmentation operations such as Gaussian smoothing, color correction, and angle adjustment are performed on the input to improve the image quality and ensure good generated results.

\textbf{Abstract representation}. In order to replace the original text information with image condition information, the image needs to be converted into embedding as input. Since the purpose of TGM is to learn the semantic information of the texture in the image and reconstruct the texture in the target generative map, rather than just a simple texture transfer and style collage. The generation result should include a high-level abstracted semantic signal and reasonable texture style, shadow parts, and fold style. To achieve this goal, we improve the model learning by compressing the conditional image information and converting it into one-dimensional conditional embedding as input. At the same time, to realize the texture control and guidance for both the upper and lower clothes, the texture of the two needs to be balanced at the junction of the clothes to achieve a reasonable texture distribution. Therefore, an averaging process is carried out when generating the input embedding to balance the texture of the upper and lower clothes. Given that the shape of the texture-guided clothes shouldn’t affect the generated results, we propose a RAT block to introduce the attention mechanism in the conditional input process to help learn high-level texture semantic information. The process of conditional input is shown in Equation (6):

\begin{small}
    \begin{equation} 
        \begin{split}
            Cond = mean(&LN(RAT_i(CLIP(T_t))), \\
    &LN(RAT_i(CLIP(D_c)))), i\in[0, n]
        \end{split}
    \end{equation}
\end{small}

\textbf{Generation prior}. In this work, we propose adopting the pre-trained model as a generation prior to providing the foundation. The benefits are twofold. First, the pre-trained model has been trained on a large amount of general data and thus can avoid implementing only a simple image collage or having limited model capabilities caused by a small training data size. Second, the adoption of a pre-trained model can significantly reduce the required model training time. Specifically, on the one hand, we use the CLIP model to extract semantic information. CLIP model has been trained on 400 million text-image pairs and can achieve good transferable performance, making it suitable for zero-shot learning and semantic extraction. On the other hand, the stable diffusion model is used as the basis of the generation. This is because the model can generate high-quality in-the-wild images, and infer reasonable characteristics in the latent space, which makes it a good foundation for texture reconstruction.

\textbf{Mask shape control}. In the try-on scene, people would like to edit the clothing area while keeping other areas as much as possible, which requires the introduction of clothing shape masks during the training phase to limit the scope of control. In order to acquire the masks, we utilize the relevant segmentation to mask the clothing area in the input. Training on this basis can help the model learn the texture information within the scope of clothing.

\subsection{Test Process}


\begin{figure}[!t]
\centering
\includegraphics[scale=0.07]{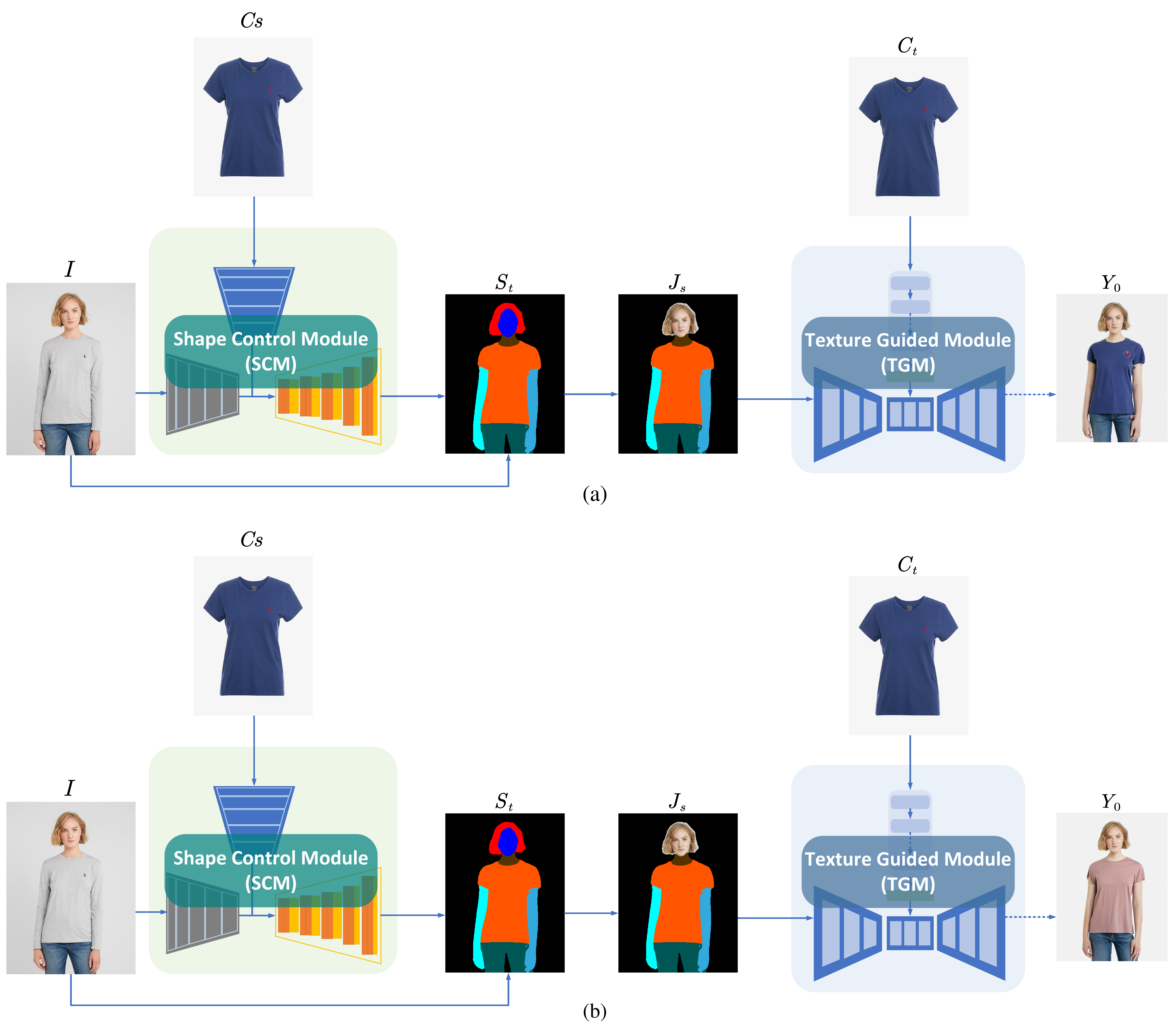}
\caption{\textbf{Test process.} The garment conditions used for shape control and texture guidance in procedure (a) are the same, which means this process fulfills the same target as the traditional virtual try-on, i.e. the target garment is worn onto the human source $I$. While in procedure (b), the garment used for shape control and texture guidance is not the same piece, which means it can realize the coupling control of the garment shape and texture.}
\label{fig: testprocess}
\end{figure}

\begin{figure*}[h]
    \centering
    \includegraphics[scale=0.19]{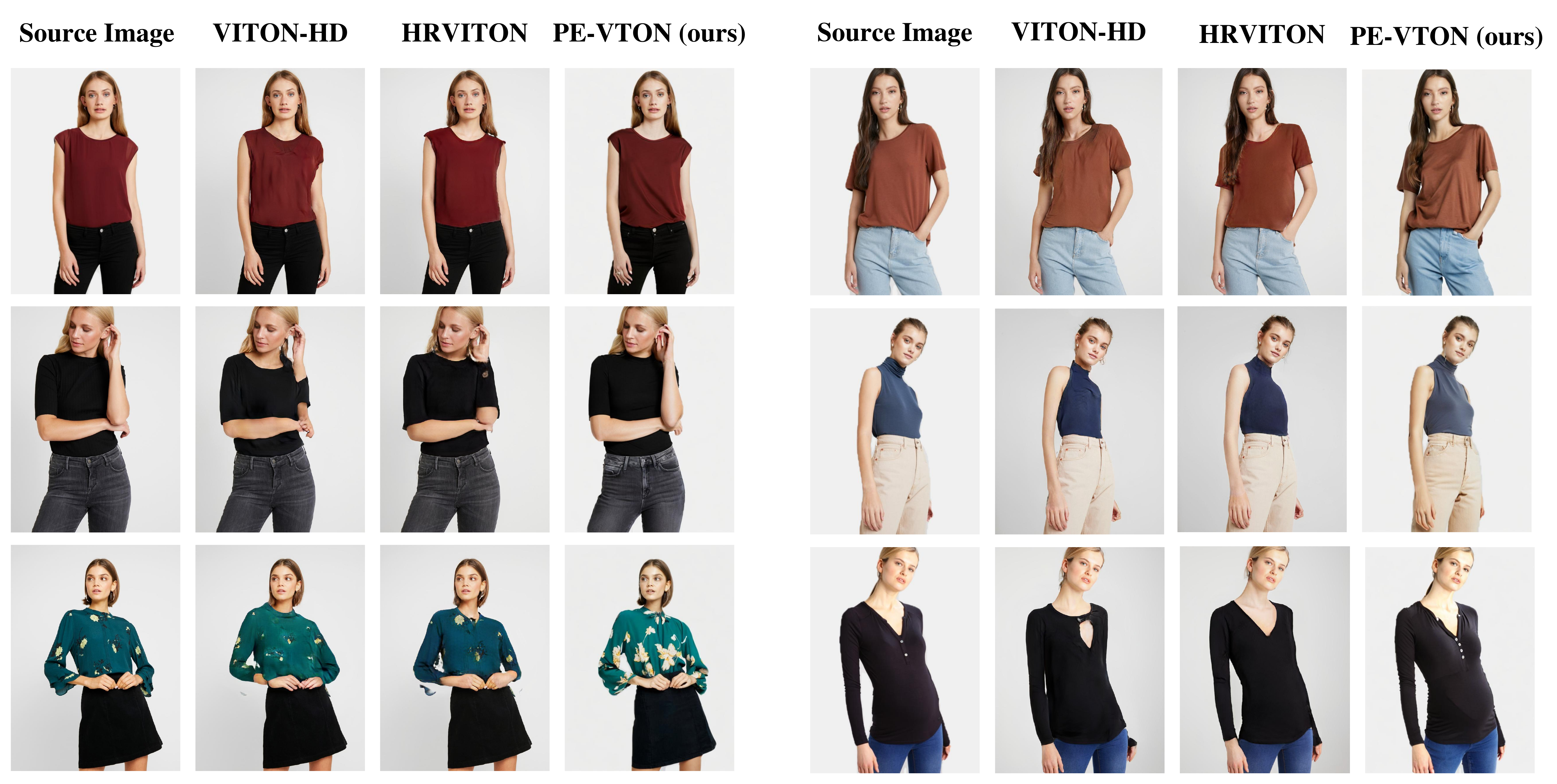}
    \captionsetup{justification=justified}
    \caption{\textbf{Qualitative comparison in VITON-HD dataset.} Compared to the results of VITON-HD and HRVITON, generative results from PE-VITON (ours) don't have artifacts and misalignment regions. Moreover, our results have a more realistic try-on effect and enhanced garment folds and texture.}
    \label{fig: qualitative comparison vitonhd}
\end{figure*}

The test process is shown in Fig. \ref{fig: testprocess}. Process (a) is the process of realizing the traditional virtual try-on, i.e., the resulting try-on result is worn in the corresponding target garment. Process (b) is the process of realizing a customized virtual try-on, which can realize separate control of the shape and texture of the garment in the generated try-on results. The inputs of SCM are the segmentation map $S_s$ corresponding to the human body image $I$ and shape conditions $C_s$. The output is the segmentation map $S_t$ whose clothing shape is the same as $C_s$. $S_t$ mainly controls the shape of the clothing through SCM. The inputs of the TGM module are the stitching image $J_s$ obtained by stitching the segmentation map $S_t$, and the human head and texture conditions $C_t$. The output is the final generation result $Y_0$, whose texture is obtained according to the texture conditions. TGM mainly guides the texture of the clothing.

Firstly, the human image $I$ is input into the SCM module for shape control. Then the control target is determined according to the clothing condition $C_s$. Finally, the segmentation map $S_t$ of the clothing shape is generated like the clothing condition $C_s$. The obtained segmentation map needs to be stitched with the human body image to obtain the stitching diagram $J_s$ that retains the human head.  $J_s$ is then sent to the TGM module for texture guidance control. The clothing clothing texture is generated as the result of clothing condition $C_t$. If the clothing condition $C_t$ and the clothing condition $C_s$ are identical at this time, the traditional virtual try-on can be realized. That is, a piece of clothing is worn on the human body. If the garment condition $C_t$ is different from the garment condition $C_s$, the final resulting garment should have the shape of condition $C_s$ and the texture style of condition $C_t$. In both cases, our models enable realistic garment folds based on the model pose and enhanced garment texture display.

\section{Experiments}

\begin{table}[!t]
\caption{\textbf{Quantitative Comparison in VITON-HD dataset and DeepFashion dataset respectively.} Compared with the paired methods and unpaired methods, PE-VITON(ours) can achieve both good result in different metrics.\label{tab:table_vitonhd}}
\centering
\scalebox{0.8}{
    \begin{tabular}{c|c|c|c|c|c|c}
         \toprule
         Dataset& Method & SSIM↑ & LPIPS↓ & IS↑ & MANIQA↑ & MUSIQ↑ \\
         \midrule
         \multirow{3}{*}{VITON-HD}
            &VITON-HD & 0.85 & 0.13 & 3.12 & 0.58 & 68.06  \\
            &HRVITON  & 0.87 & 0.12 & 3.13 & 0.61 & 67.93\\
            & \textbf{Ours} & \textbf{0.92} & \textbf{0.07} & \textbf{3.24} & \textbf{0.71} &\textbf{78.20}\\
         \midrule
         \multirow{3}{*}{UPT-512}
            & PASTA-GAN & 0.88 & 0.54 & 1.33 & 0.59 & 65.83 \\
            & PASTA-GAN++ & 0.87 & 0.59 & 1.30 & 0.61 & 69.15 \\
            & \textbf{Ours} & \textbf{0.94} & \textbf{0.04} & \textbf{3.49} & \textbf{0.73} & \textbf{78.88} \\
         \bottomrule
    \end{tabular}
    }
\end{table}

\begin{table}[!t]
\caption{\textbf{Quantitative Comparison of the proposed model trained on different datasets.} Deep-M represents the DeepFashion-Multimodal dataset. This table shows the results trained in different models, which are trained with different datasets.\label{tab:table_deepfashion}}
\centering
\scalebox{0.87}{
    \begin{tabular}{c|c|c|c|c|c}
         \toprule
        Datasets & SSIM↑ & LPIPS↓ & IS↑ & MANIQA↑ & MUSIQ↑ \\
        \midrule
        VITON-HD & 0.85 & 0.45 & 1.50 & 0.50 & 67.46 \\
        \midrule
        Deep-M & 0.85 & 0.45 & 1.45 & 0.41 & 62.77 \\
        \midrule
        VITON-HD + Deep-M & 0.85 & 0.45 & 1.51 & 0.47 & 65.69 \\
        \bottomrule 
    \end{tabular}}
\end{table}

\begin{figure*}[h]
    \centering
    \includegraphics[scale=0.195]{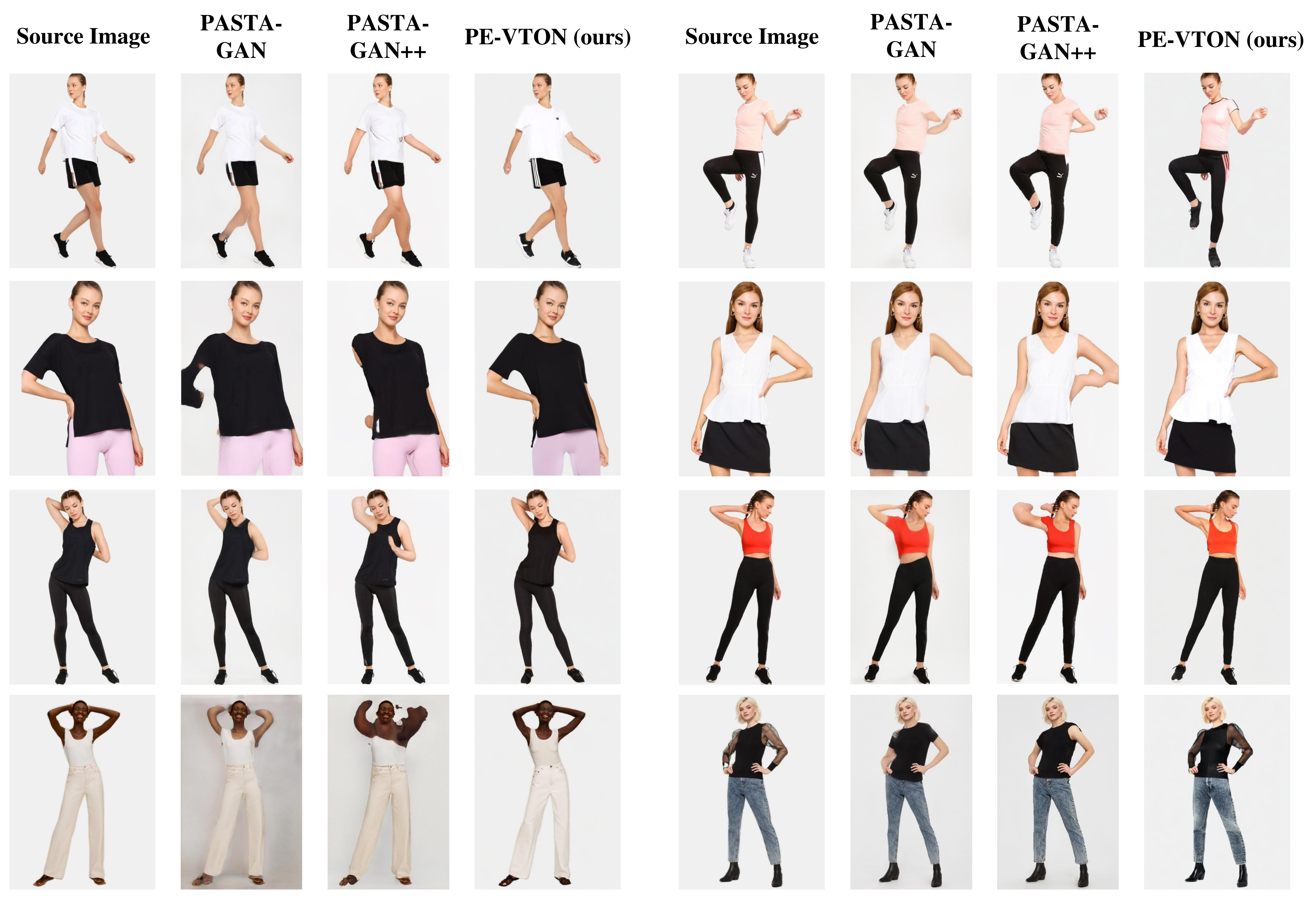}
    \captionsetup{justification=justified}
    \caption{\textbf{Qualitative comparison in UPT dataset.} Compared to the results of PASTA-GAN and PASTA-GAN++, Using PE-VITON (ours) can generate more rational limbs and better deal with complex poses and clothing details.}
    \label{fig: qualitative comparison upt}
\end{figure*}

\subsection{Experiment Setup}

\textbf{Datasets}. We conduct experiments on the Deepfashion-Multimodal and VITON-HD datasets to verify the effectiveness of the proposed method. The Deepfashion-Multimodal dataset contains 12701 full-body images in 24 categories, with the resolution of 751×1101. The VITON-HD dataset contains 13679 pairs of half-body view and upper in-store clothing in 13 categories, with the resolution of 1024×768. Due to the inconsistent category labels between the two datasets, we modify the category labels of DeepFashion-Multimodal dataset based on the labels used in VITON-HD dataset. Next, we merge the data from both datasets and performed a redivision to create new training and testing sets, where the training set includes 11647 pairs and 39685 independent images; and the testing set includes 2032 pairs and 4410 independent images.

\textbf{Implementation details}. The model is implemented using 2 NVIDIA 3090Ti GPUs with the batchsize setting of 8. We used the Adam optimizer as the gradient descent method and set the learning rate as 1.0e-05. The model adoptes the iterative training method, through which the model is first trained at 192 resolution, and gradually adjusted till the final resolution reached 768.

\textbf{Metrics}. Our goal is to apply the target clothing shape and texture to the source person image so that the edited parts are similar to the target clothing and the generated result is realistic. To measure the model quality comprehensively, we evaluate the generated images by adopting the following 5 metrics. 1) Multi-Scale Structural Similarity (MS-SSIM) is used to evaluate structural similarity between two images. 2) Learned Perceptual Image Patch Similarity (LPIPS) is used to calculate the distance between images generated in the perception domain and ground truths. 3) Inception Score (IS) is used to reflect the clarity and diversity of the generated image. 4) Multi-dimension Attention Network for no-reference Image Quality Assessment (MANIQA) is used to evaluate the reality and clarity of the generated images. 5) MUlti-Scale Image Quality transformer (MUSIQ) is used to evaluate the quality and clarity of the generated images.

\subsection{Performance Comparison of Different Algorithms}

\begin{figure*}[h]
    \centering
    \includegraphics[scale=0.18]{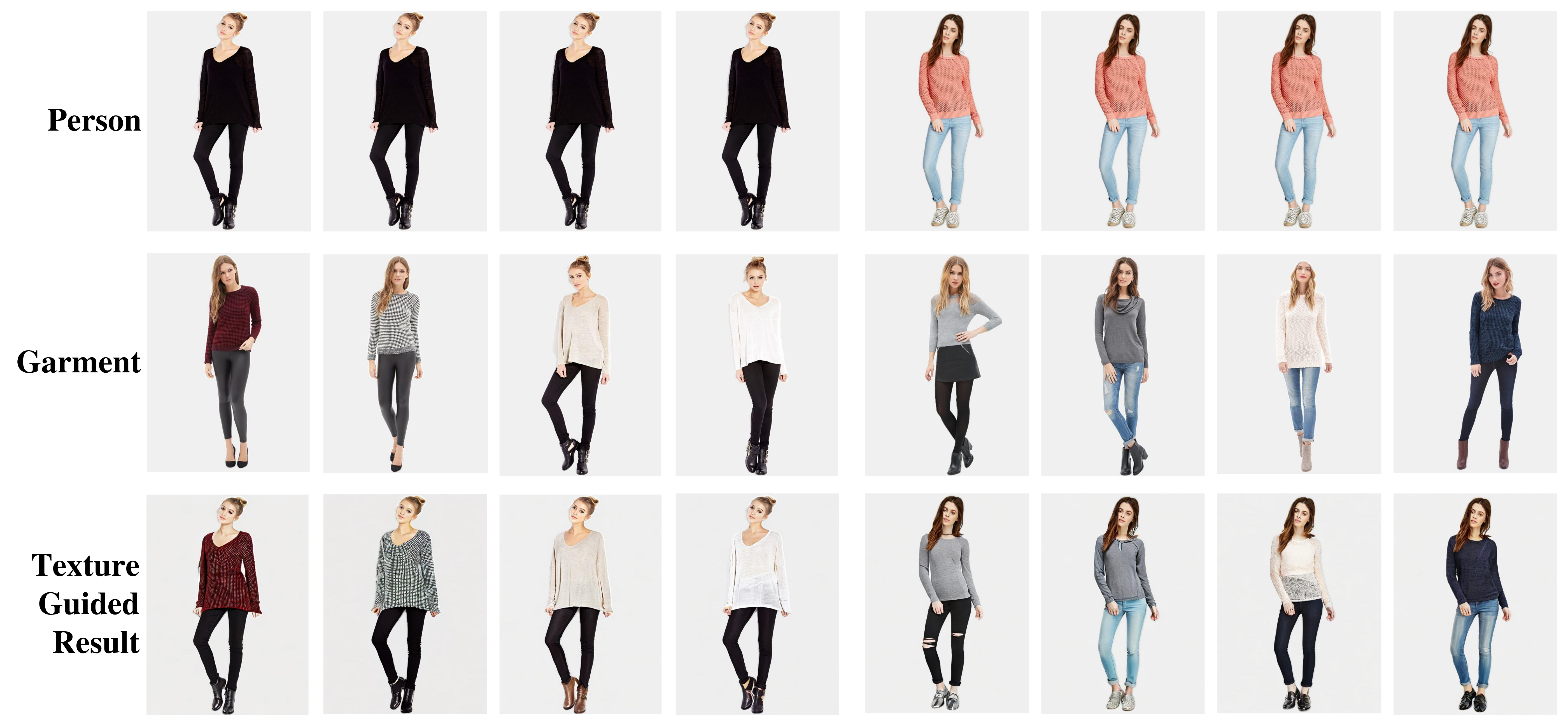}
    \captionsetup{justification=justified}
    \caption{\textbf{Texture guidance.} Using a conditional garment to change the texture of the source image cloth, so that the resulting result has the texture of the conditional garment and the shape of the source image.}
    \label{fig: texture guidance}
\end{figure*}

\textbf{Qualitative Comparison.}We compare our method with several state-of-the-art virtual try-on methods, including VITON-HD, HR-VITON, PASTA-GAN and PASTA-GAN++. Fig. \ref{fig: qualitative comparison vitonhd} shows the visual comparisons of VITON-HD, and HR-VITON on the VITON-HD datasets. Fig. \ref{fig: qualitative comparison upt} shows the visual comparisons of PASTA-GAN and PASTA-GAN++ on the UPT-512 dataset. It can be seen that compared with these methods, the results obtained by our model can better retrain human body information, handle complex postures, generate more realistic texture styles and obtain realistic clothing folds.

As shown in Fig. \ref{fig: qualitative comparison vitonhd}, compared to VITON-HD and HR-VITON, our method is able to not only generate more natural body shape and body edge but also distinguish the area of clothing and arm well. VITON-HD tends to generate unnatural garment folds and blurred garment textures, making it difficult to reveal the details of garments. HR-VITON tends to smooth the texture of the garment, leading to a reduction in the authenticity of the generated result. Additionally, VITON-HD and HR-VITON have not eliminated the artifacts in the misaligned regions completely, which leads to the unexpected obstruction of the torso by the garment, as well as the deformation of the generated arms. Through the SCM module, the clothing can be better aligned with the human body part, so as to generate a more accurate and clearer segmentation map. Furthermore, the diffusion model helps provide texture and folds enhancements. As can be seen from the upper left and lower left figures in Fig. \ref{fig: qualitative comparison vitonhd}, our generated results further enhance garment folds and styles, which are generated according to the human posture from the source image and the texture from the target try-on garment, making the results more realistic and ornamental. 

As can be seen from Fig. \ref{fig: qualitative comparison upt}, PASTA-GAN and PASTA-GAN++ are difficult to generate real limbs, especially when the human body's postures are complex. Moreover, since these methods split clothing into small patches, this can easily lead to clothing misalignment at the connection of different patches. When the garment is split into patches, the details are lost, which causes the generated result to lack garment details and reduces the authenticity of the result.


\begin{figure}[!t]
\centering
\includegraphics[scale=0.19]{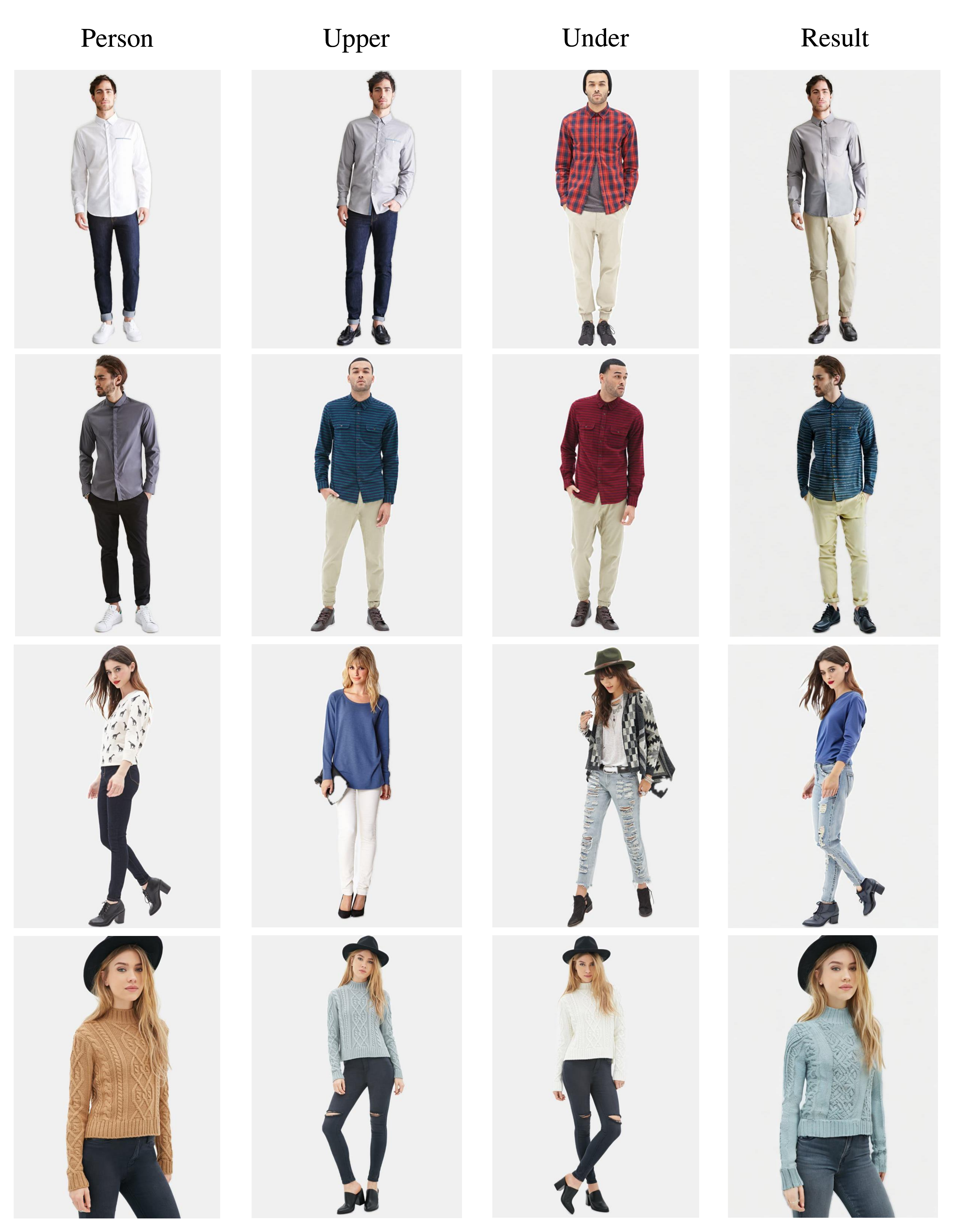}
\caption{\textbf{Full body outfit.} The top garment and the bottom garment are derived from different images.}
\label{fig: full body outfit}
\end{figure}

\textbf{Quantitative Comparison.}
We perform quantitative experiments on VITON-HD datasets and UPT-512 datasets respectively. Because PASTA-GAN and PASTA-GAN++ are used for the resolution of 320×512 at most, while the DeepFashion-Multimodal dataset owns the resolution of 750×1101. To achieve a fair comparison, we use the UPT-512 dataset for quantitative and qualitative comparison rather than the DeepFashion-Multimodal dataset. Table \ref{tab:table_vitonhd} lists the scores of our method on the HRVITON dataset and the UPT dataset respectively, as well as the results compared to the state-of-the-art high-resolution try-on method. As can be seen from the tables, our method performs significantly better than the comparison methods, showing that the proposed method achieves high-quality, high-definition, and high-fidelity try-on results on both paired and unpaired datasets.

Our model is trained simultaneously on two datasets, VITON-HD and DeepFashion-Multimodal. To prove that the generative ability of the model is universal, we train submodels on the VITON-HD dataset and DeepFashion-Multimodal dataset respectively, and compare them with our model. To achieve a fair comparison, we collect pictures of the wild for testing. Table \ref{tab:table_deepfashion} lists the scores of different submodels. It can be seen that all submodels can achieve good results, indicating our model owns good learning ability rather than simply relying on the dataset.

\subsection{Dress-up effect of PE-VITON}
Since the shape and texture of clothing are decoupled, the specific properties of clothing can be directly controlled to change through asynchronous operations, so as to realize a series of applications such as clothing editing.

\textbf{Texture guidance}. As shown in Fig. \ref{fig: texture guidance}, our model can guide the current result to generate a similar texture style by entering the corresponding texture style or other clothing, and further optimize the generated result by enhancing the texture during the texture migration.

\begin{figure}[h]
    \centering
    \includegraphics[scale=0.12]{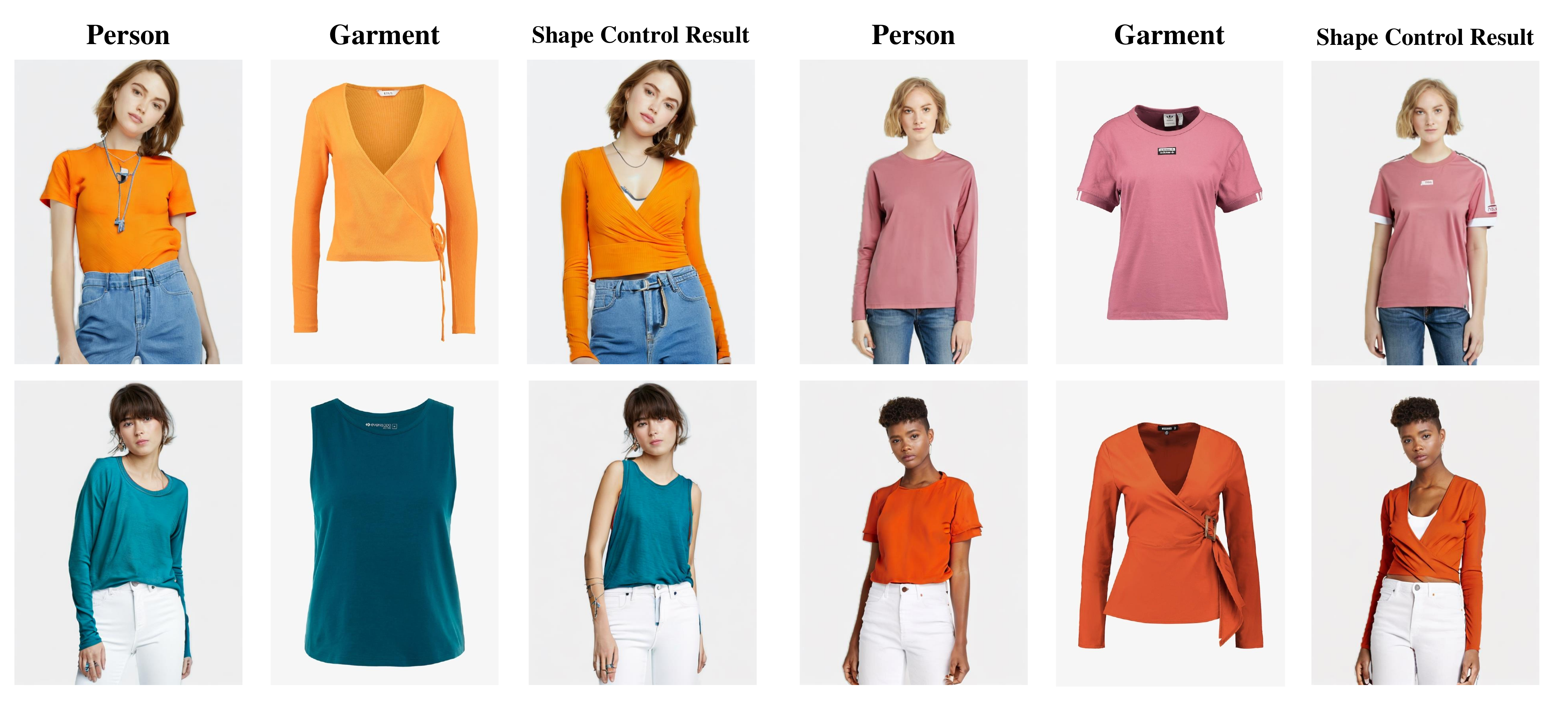}
    \captionsetup{justification=justified}
    \caption{\textbf{Shape control.} Using the conditional garment to change the shape of the source image cloth, so that the generative result has the shape of the conditional garment and the garment texture of the source image.}
    \label{fig: shape control}
\end{figure}

\textbf{Full body outfit}. As shown in Fig. \ref{fig: full body outfit}, our model allows tops to be freely matched with bottoms. By integrating clothing styles from different human bodies, so as to achieve personalized fitting of tops and bottoms on one body. We need to achieve this application with two forward passes. First, only the top garment is changed, and the SCM module generates the segment with the target top garment only. Next, change the bottom garment with a similar process. It can be seen that our model is able to generate reasonable clothing folds according to different human postures.

\textbf{Shape control}. As shown in Fig. \ref{fig: shape control}, our model enables free change of clothing shape by changing only the shape of the garment and retaining the current garment texture. This application offers a new way of approach to clothing editing, that allows them to control the generated results by modifying the clothing segmentation.

\section{Conclusion}
In this paper, we propose a novel personalized virtual try-on solution with shape control and texture (PE-VITON) that decouples garment shapes and textures to enable personalization and modification. In the solution, the Shape Control Module (SCM) and the Texture Guidance Module (TGM) are proposed to change the shape and the garment texture respectively. The proposed SCM can solve the problem of artifacts caused by garment-human misalignment. The proposed TGM can generate real garment folds and restore natural human body parts. A large number of experiments have proved that the PE-VITON has the ability to handle a variety of clothing, and its try-on effect is superior to other advanced methods in paired and unpaired datasets. We believe this method has some inspiration for the directional generation task of diffusion model, and further expands the application scenarios of virtual try-on.

\section{Limitation and Future Work}
The conditional input of our method is the embedding with the cloth features generated by CLIP. Due to the diversity and uncertainty of clothing patterns, it is difficult for the generated embedding to include all details of clothing, which leads to the subtle change of clothing textures when the model generates the try-on results. Although our method currently has a wider range of application scenarios and stronger adaptability than other methods, there is still room for improvement. In the future, we plan to optimize the model's texture generation capabilities with more accurate control.


\bibliographystyle{ieeetr}
\bibliography{references}

\begin{IEEEbiography}
[{\includegraphics[width=1in,height=1.25in,clip,keepaspectratio]{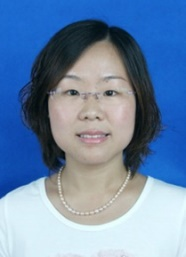}}]
{Shufang Zhang}
is currently an associate professor in the School of Electronics and Information Engineering, Tianjin University, Tianjin, China. She received the M.s. degree and the Ph.D. degree from Tianjin University in 2004 and 2007 respectively. Her research interests include deep learning, Virtualtry-on, and outfit recommendation.
\end{IEEEbiography}

\begin{IEEEbiography}
[{\includegraphics[width=1in,height=1.25in,clip,keepaspectratio]{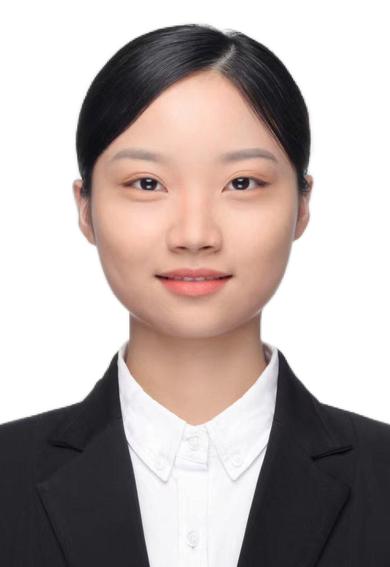}}]
{Minxue Ni}
received B.S. degree in the school of Electrical and Information
Engineering from Tianjin University in 2022, Tianjin, China. She is currently working toward her Master degree in the School of Electrical and Information Engineering, Tianjin University, Tianjin, China. Her research interests mainly focus on computer version for virtual try-on.
\end{IEEEbiography}

\begin{IEEEbiography}
[{\includegraphics[width=1in,height=1.25in,clip,keepaspectratio]{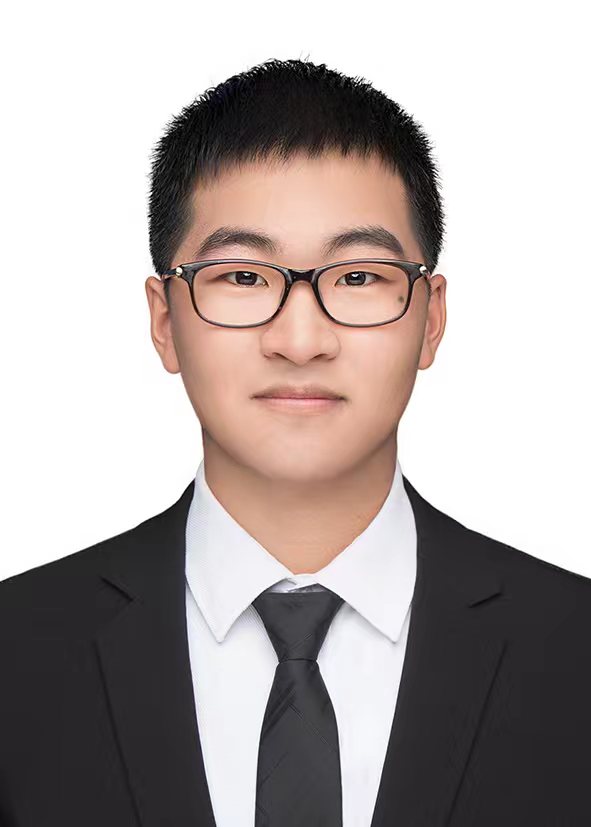}}]
{Lei Wang}
received B.S. degree in 2022 from Tianjin University, Tianjin, China. He is currently working toward the Master degree in Tianjin University. His current research includes Embedded development and virtual try-on.
\end{IEEEbiography}

\begin{IEEEbiography}
[{\includegraphics[width=1in,height=1.25in,clip,keepaspectratio]{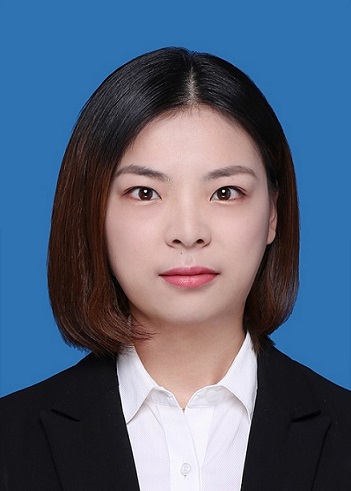}}]
{Wenxin Ding}
received the B.S. degree and M.S. degree in the School of Electrical and Information Engineering from Tianjin University in 2016 and 2019, respectively. She is currently pursuing her Ph.D. in the School of Electrical and Information Engineering, Tianjin University, Tianjin, China. Her research is mainly about outfit recommendation and Virtual try-on.
\end{IEEEbiography}

\newpage

\begin{IEEEbiography}
[{\includegraphics[width=1in,height=1.25in,clip,keepaspectratio]{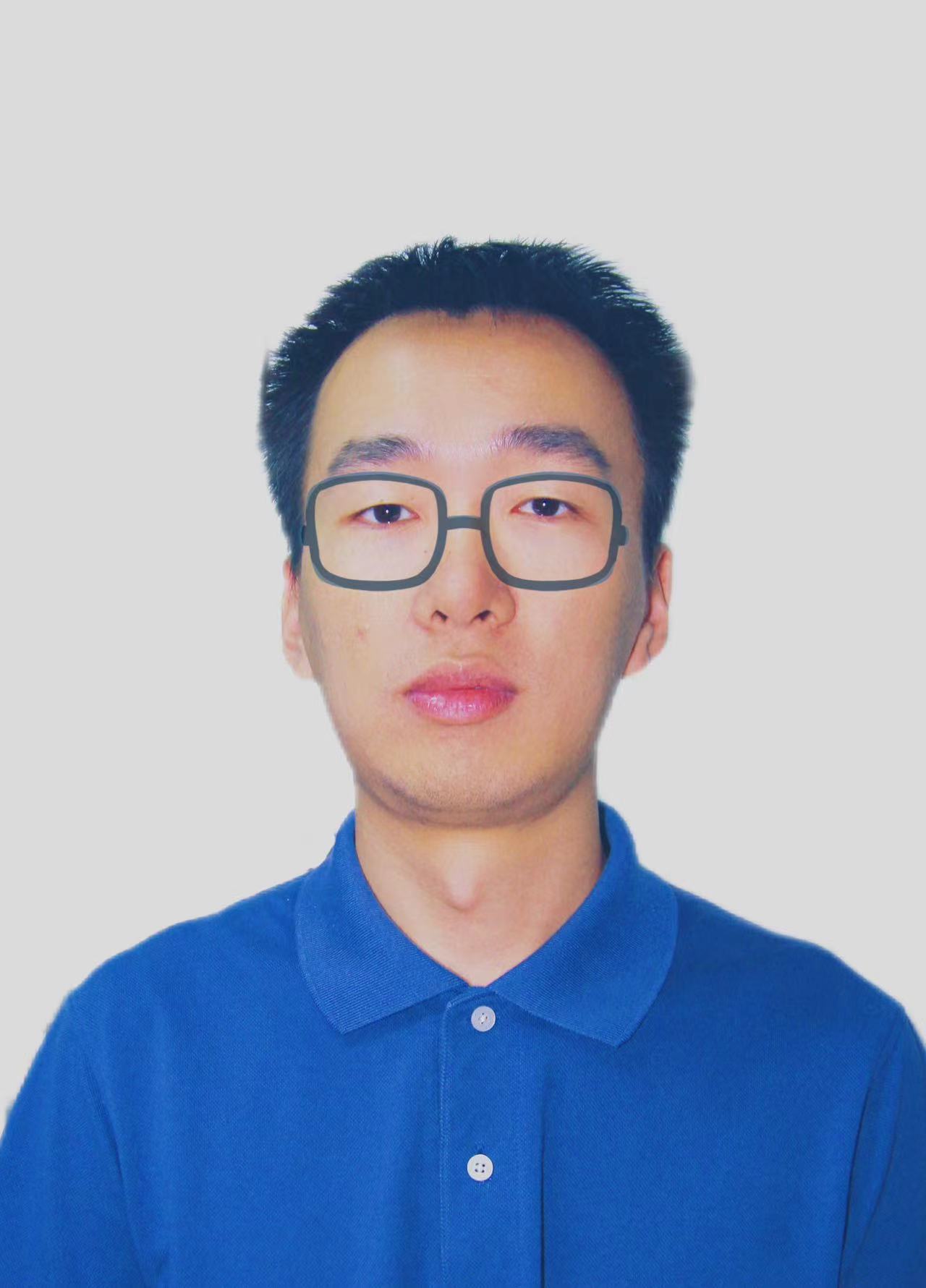}}]
{Shuai Chen}
is a senior Aritficial Intelligence algorithm engineer for Hisense. He obtained a bachelor's degree from Ocean University of China. He is currently working on large-scale language models and image generation. He is currently engaged in AIGC-related work.
\end{IEEEbiography}

\newpage

\begin{IEEEbiography}
[{\includegraphics[width=1in,height=1.25in,clip,keepaspectratio]{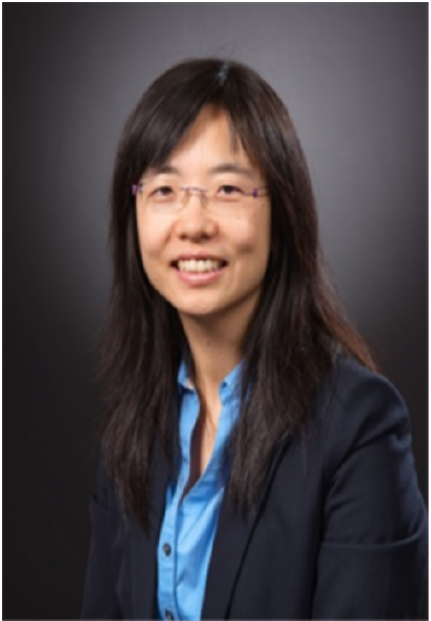}}]
{Yuhong Liu}
received a Ph.D. degree from the University of Rhode Island in 2012. She is the recipient of the 2013 University of Rhode Island Graduate School Excellence in Doctoral Research Award. She is an associate Professor at Department of Computer Science and Engineering from Santa Clara University. Her research interests include trustworthy computing and cyber security of emerging applications.
\end{IEEEbiography}



\vfill

\end{document}